# AI vs Human Expert Reasoning:

## Assessing Agreements in Building Typology Predictions based on Street View Imagery

Zahratu Shabrina[1*], Muhammad Asa[2], Jin Rui[1], Lu Yin[3], Stephen Law[4]

[1] Department of Geography, King's College London, UK
[2] Resilience Development Initiative, Indonesia
[3] School of Computer Science and Electronic Engineering, University of Surrey, UK
[4] Department of Geography, University College London, UK
*_Corresponding author_

**Abstract**

This research investigates the potential of Vision-Language Models (VLMs) to infer building typologies, *Construction*, *Current Use*, *and Storeys* from Google Street View (GSV) images. Predictions generated by VLMs are compared with inference by human experts (civil engineers and architects) as a source of manually labelled ground-truth data. We evaluate several state-of-the-art VLMs, including GPT-4o, Claude 3.5 Sonnet, and Gemini 2.0 Flash. By applying different scaling strategies and prompting techniques, we found that Chain-of-Thought prompts provide an overall more stable model performance. We also investigate the reasoning behind VLMs' building-typology predictions by examining the probabilities of keywords appearing in AI explanations. This enabled us to analyse patterns in these reasonings and identify key themes driving both agreements and disagreements between VLM and expert labels. We find that AI tends to focus on visual indicators, whereas human experts place greater emphasis on broader contextual cues and domain knowledge, in addition to visual cues. Overall, VLM can approximate experts' capability in building-typology classification at scale, with an average accuracy of approximately 70%. The study demonstrates the VLM's potential for AI automation in tasks that require pattern recognition and object identification in an urban context. AI have the potential to serve as complementary and collaborative tools for urban analysis, leveraging their strengths in understanding visual patterns. This study contributes to the exploration of the efficiency and scalability of AI visual prediction and provides insights into the reasoning processes that could support automation processes in urban analysis and prediction.

## 1. Background

Comprehensive building-level data is an important source for understanding, analysing, and simulating urban problems to support more effective and sustainable planning and policy development. It has been used to understand urban fabric classifications (Perez et al., 2018), energy modelling (Cerezo et al., 2017; Reinhart & Cerezo Davila, 2016; Sakti et al., 2022), and vulnerabilities to hazard (Blagojević et al., 2023; Lang et al., 2018; Nofal et al., 2024), among many potential uses. Predictions of building typologies, such as buildings' locations, ages, structures, number of storeys, and current use, can enable more robust and accurate computational analysis and modelling by providing data at granular spatial resolution. This is particularly useful when macro-urban-scale models have limited capacity for policymakers seeking to provide practical, targeted policy interventions. However, information at fine spatial scales, including building typologies, is often outdated, spatially fragmented, incomplete, and heterogeneous (Hecht et al., 2015). This is especially prevalent in the Global South, where data collection frequently relies on traditional surveys and assessments that are often costly, time-consuming, and difficult to scale.

The study leverages the robust Google Street View (GSV) dataset, which provides a detailed visual representation of the urban environment through panoramic, street-level imagery captured by vehicle-mounted cameras (Anguelov et al., 2010). Throughout the years, street view imagery has enabled large-scale analysis in various disciplines, such as for estimating house prices (Law et al., 2019; G. Y. Wang, 2023; Y. Zhang & Dong, 2018), analysing travel and mobility patterns (Goel et al., 2018; S et al., 2019; L. Yin & Wang, 2016), as well as investigating the relationship between health and built environment (Keralis et al., 2020; Nguyen et al., 2021). We examine the potential of technological advancements, including Vision-Language Models (VLMs), to predict multiple building-level data from street-view imagery in the Global South, using Jakarta, Indonesia, as a case study. We draw on recent advances in Vision-Language Models (VLMs), a class of Artificial Intelligence (AI) systems that can *see* and *understand* visual and textual information through deep learning. VLMs such as GPT-4o, Claude 3.5 Sonnet, and Gemini 2.0 Flash extend the capabilities of Large Language Models (LLMs) by incorporating a visual encoder and an adapter that aligns image features with text embeddings, thereby enabling the joint processing of multimodal inputs (Kaduri et al., 2024). This architecture allows VLMs not only to generate text about images but also to classify, compare, and reason across modalities in ways that mimic human interpretation.

We focus on Penjaringan, a northern coastal district of Jakarta, Indonesia, one of the city's poorest communities. The choice of the study area is intentional, as northern coastal Jakarta offers a balanced mix of building typologies, including low- to high-rise buildings and formal and informal settlements, with a variety of building types and an incomplete GSV database. This offers a typical model of the Global South's major cities, providing a generalisable testbed for similar cities, particularly in Southeast Asia. The North Jakarta area faces overlapping socio-environmental challenges, including speculative real estate development, large-scale land reclamation, land subsidence, and recurrent tidal flooding (Colven, 2020b). In such contexts, where informal settlements are prevalent, existing datasets often fail to capture the fine-grained characteristics of the built environment. By predicting building-level data, it becomes possible to generate a high-resolution urban analysis that incorporates critical factors such as construction materials and current building use. This approach is particularly valuable in areas such as Penjaringan, where housing is typically self-built, incrementally expanded, and adapted through ad hoc measures (Karunaratne, 2023). These forms of organic, bottom-up construction are rarely reflected in official records or neighbourhood-scale surveys. Yet they play a central role in shaping community characteristics and associated vulnerabilities.

This study offers several contributions to address the knowledge gap in VLM research. First, there are still limited studies that have used GSV to infer multiple building-level data, and previous studies often focus on a single typology, such as building type classification (Hoffmann et al., 2019), building materials (Firmansyah et al., 2024; Gonzalez et al., 2020) and building age (Sun et al., 2021). Our study extends the approach to predicting the multi-dimensional nature of the building stock, enabling predictions for multiple datasets: *Constructions, Current Use, and Storeys* attributes. Second, despite their rapid adoption, VLMs are still treated primarily as "black boxes," with limited scholarship examining the mechanisms underlying their predictions. Our study examines both the predictive outputs and the reasoning processes of VLMs, comparing them with expert judgements, which gives nuance to the reasoning behind closed-source VLM for image prediction. By analysing where VLMs align with or diverge from expert reasoning, we aim to provide deeper insights into how these models operate and how their outputs can be interpreted and potentially used in applied urban research.

Third, this paper presents a case study from a diverse, heterogeneous, data-scarce perspective in the Global South. As commercial VLMs are mainly trained on the Global North dataset, they often exhibit Western-centric biases from high-resource countries (Ananthram et al., 2025). This study allows for some degrees of interpretability in understanding VLM predictions using the Global South dataset. Fourth, the studies on building stock and typology predictions are predominantly reliant on neural networks and deep learning trained on labelled images, making this paper among the first to leverage VLMs to enable semantically grounded predictions without requiring task-specific architecture. Lastly, this study explores different types of prompting to examine differences in prediction results based on various prompt strategies.

## 2. Literature Review

### 2.1. Street View Imageries as a Source of Visual (Big) Data

The importance of street-view imagery as an urban data source for analysing urban perception and infrastructure has been widely recognised in multidisciplinary research (Biljecki & Ito, 2021). Street View is a source of easily accessible, large-scale (imagery) data that is both cost- and time-effective for collecting visual datasets for various research purposes, with Google as the main data source due to its reliability and image quality (Rzotkiewicz et al., 2018). Street View imagery allows researchers to view a place from the eyepoint rather than the conventional birds-eye view, unlike satellite imagery. Based on a systematic review of 619 papers, Google Street View (GSV) currently dominates research on street-view imagery, ranging from urban perception studies, spatial data infrastructure, transportation, health and wellbeing, greenery and socioeconomic studies (Biljecki & Ito, 2021). For instance, Law et al. (2019) used deep neural networks to extract features from London street-view and satellite imagery, finding that incorporating visual information significantly improved house price predictions. Nguyen et al., (2021) analysed street-view images covering 2,916 counties in the U.S., constructing a spatialized indicator system for built environment features to systematically evaluate their statistical associations with U.S. county-level resident health outcomes. Others have utilised GSV for criminology research (Connealy & Corts, 2024; Sundquist et al., 2025; Sytsma et al., 2021; Vandeviver, 2014), as well as climate change research, such as surface temperature prediction(Klimenka et al., 2025; T. Zhang et al., 2024), among many other research strands.

Despite its potential, using GSV for analysis has its own challenges that need to be acknowledged when using it as a data source. Google provides a proprietary dataset that comes with its own biases and terms and conditions. GSV spatial coverage is highly skewed in high-income countries with adequate road infrastructure quality. This is mainly due to low demand and challenges posed by GSV survey vehicles in accessing densely populated informal urban settlements in lower-income countries (Rzotkiewicz et al., 2018; Umar et al., 2023). The incomplete spatial coverage introduces potential biases, such as underestimation of urban phenomena (e.g., waste densities), which require adjustments to study metrics (Umar et al., 2023). Fan et al., (2025) pointed out that Google Street View imagery also has systematic collection biases in urban environment mapping, covering only about 62.4% of building facades and tending to over-represent non-residential buildings, which may lead to degraded performance of AI models trained on such imagery in marginalised communities. Other challenges include image quality, the frequency of image capture, and the potential for some images to be blocked or blurred in certain cases.

### 2.2. Technological Frontiers in Vision-Language Models and Prompting Strategies

The development of vision-language models (VLMs) marks a significant transition from task-specific supervised learning in computer vision to a more general, multimodal understanding. Previously, most vision-language tasks (e.g., object detection, image classification, and semantic segmentation) relied on large-scale, task-specific, crowd-labelled data and deep neural network (DNN) training, which is often time-consuming and laborious (J. Zhang et al., 2024). Nowadays, there is an increasing interest in zero-shot learning, the ability of AI to perform tasks and recognise objects without seeing them in the training data. A VLM is pre-trained on large-scale data available on the internet, and its analysis can be directly applied without fine-tuning, such as with the CLIP model proposed by Radford et al. (2021). The model achieved transferable visual representations through contrastive learning pretraining on 400 million image-text pairs, with performance comparable to that of the ResNet-50 model on zero-shot ImageNet classification, demonstrating strong potential for zero-shot transfer across a wide range of tasks (Radford et al., 2021).

With the emergence of generative AI, commercialised VLMs such as GPT-4V, Claude 3.5 Sonnet, and Gemini 2.0 Flash deeply integrate the reasoning capabilities of large language models with visual understanding. Liu et al. (2023) introduced the LLaVA model via instruction tuning, combining a visual encoder with an LLM to achieve general visual-language understanding, achieving an accuracy of 92.53%. However, Huang et al. (2025) reported "hallucination" errors in VLMs during complex visual reasoning. These systematic inaccuracies highlight ongoing challenges and concerns for high-stakes applications where precision and reliability are critical.

In zero-shot scenarios, a model's success often hinges on how well the prompt aligns with its internal representations and the objectives of its training data. This is known as prompt engineering/strategies, a core mechanism for guiding large model behaviour that has its theoretical foundation rooted in the "in-context learning" concept proposed by Brown et al. (2020). Z. Yin & Wang, (2025) proposed type-guided Chain-of-Thought (CoT) prompting, which adds an additional reasoning step in scientific table understanding, significantly improving model accuracy on multi-step reasoning tasks. In the visual task domain, Xiao et al. (2025) showed that decomposition prompting has advantages for single complex tasks and optimises answer quality through decomposition, reflection, and correction mechanisms. However, research on prompting strategies remains limited. Most studies focus on fine-tuning and evaluation of open-source models, yet they lack systematic comparisons of low-threshold, easily deployable, commercialised VLMs. The selection of prompting strategies is often based on experience rather than systematic experiments, and there is a lack of understanding of the performance boundaries of zero-shot, Chain-of-Thought, single-task, and multi-task prompting in specific application scenarios. Turpin et al. (2023) further revealed that the causal relationship between CoT-generated reasoning steps and final answers may be "post-hoc rationalisation," posing fundamental questions for interpretability research in VLMs.

Although recent advances have established a strong methodological foundation for understanding Vision Language Models (VLMs) and their capacity for visual perception modelling, several key gaps remain in applying these models to predict building typologies. First, existing studies have not systematically compared the performance of *commercial* VLMs or evaluated prompt design strategies for predicting building typology. While many investigations have explored VLMs in visual reasoning tasks, they often rely on fine-tuned open-source models that require substantial computational resources and are impractical for many applied urban studies. Commercial VLMs, such as GPT-4o**,** Claude 3.5 Sonnet, and Gemini 2.0 Flash, offer zero-shot inference capabilities that reduce the technical and computing barriers to use. However, there is limited evidence on how these models perform across different prompts or typology prediction tasks, particularly in applied, real-world urban contexts.

Second, the reasoning processes of VLMs remain underexplored. Most existing studies treat these models as 'black box' systems, evaluating their outputs primarily on prediction accuracy without examining how they arrive at these results or predictions. Although explainable AI methods such as attention visualisation have been employed, they provide limited insight into how VLMs translate visual features into semantic reasoning. To address this, our study requires VLMs to produce structured 200-word reasoning explanations for each prediction, which are then analysed using text analysis. This enables a systematic comparison of machine reasoning and expert interpretation, offering insights into the cognitive alignment and divergence between them. Third, there is a notable lack of research applying VLMs to areas characterise with strong presence of informal settlements in the Global South. Much of the existing work focuses on formal urban contexts in North America and Europe, where data availability and urban form differ substantially. In contrast, cities in the Global South are characterised by rapid informal urbanisation, limited or outdated datasets, and heterogeneous, self-built environments (Karunaratne, 2023). These conditions pose challenges for pre-trained models that may embed geographic and socioeconomic biases.

## 3. Data and Methods

### 3.1. Data and Study Area

This study utilises three primary sources of data: (1) building-level footprints from OpenStreetMap, (2) Google Street View (GSV) images across Penjaringan and (3) expert-labelled ground truth data. OpenStreetMap, an open dataset collected through crowdsourcing[1], was selected as a building footprint data source because it provides the most accurate representation of the area. The centroid coordinates of each building were derived as input for Google's Street View (GSV) Static API. The API enabled us to automatically adjust the camera angle to directly face each building by supplying the corresponding coordinates. The API's other parameters (e.g., field of view, pitch, size, radius, source) were adjusted according to the characteristics of the available Google Street View imagery in Penjaringan.

[1] The official building-level data by the government was deemed outdated as it was based on an aerial mapping in 2016.

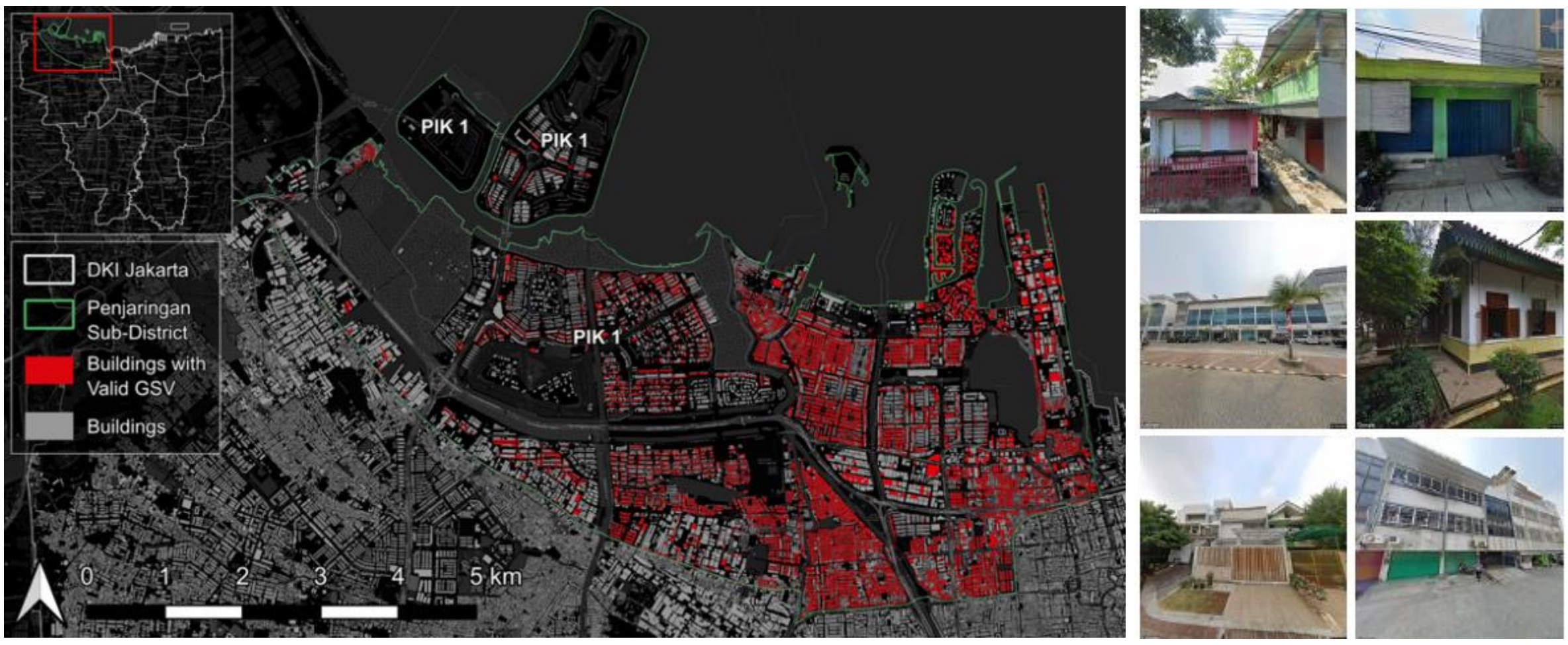

*Fig 1. The Locations of Buildings with Available and Valid GSV data in Penjaringan District, North Jakarta*

A total of 63,729 street-view images were collected from the API, of which 29,937 were considered valid for predictions after a series of preprocessing steps, including data screening using LLaVA-v1.6-Mistral-7B. The locations of buildings with available and valid GSV data are shown in Fig 1, denoted in red. The filtering process involves three steps. First, we removed invalid images, characterised by their low file sizes. Second, buildings not visible from the street are excluded to increase spatial accuracy. Those buildings were identified by inverse selection of buildings intersected by a 10-m buffer polygon of the road network dataset from the Government of Jakarta, representing the 95$^{th}$ percentile of Jakarta's road width. Third, we utilised the LLaVA model to automatically remove images in which the building is significantly occluded by foreground objects (e.g., trees, vehicles, pedestrians).

### 3.2. Methods

This study examines both the predictive outputs and the reasoning processes of Vision-Language Models (VLMs), comparing them with expert judgements to assess reliability and interpretability in urban research. We adopt a stepwise, scalable approach using Google Street View (GSV) imagery. The zero-shot building prediction was conducted using popular commercial pre-trained vision language models (VLMs), such as OpenAI GPT-4, Anthropic Claude 3.5 Sonnet, and Google Gemini 2.0 Flash, as shown in Fig 2.

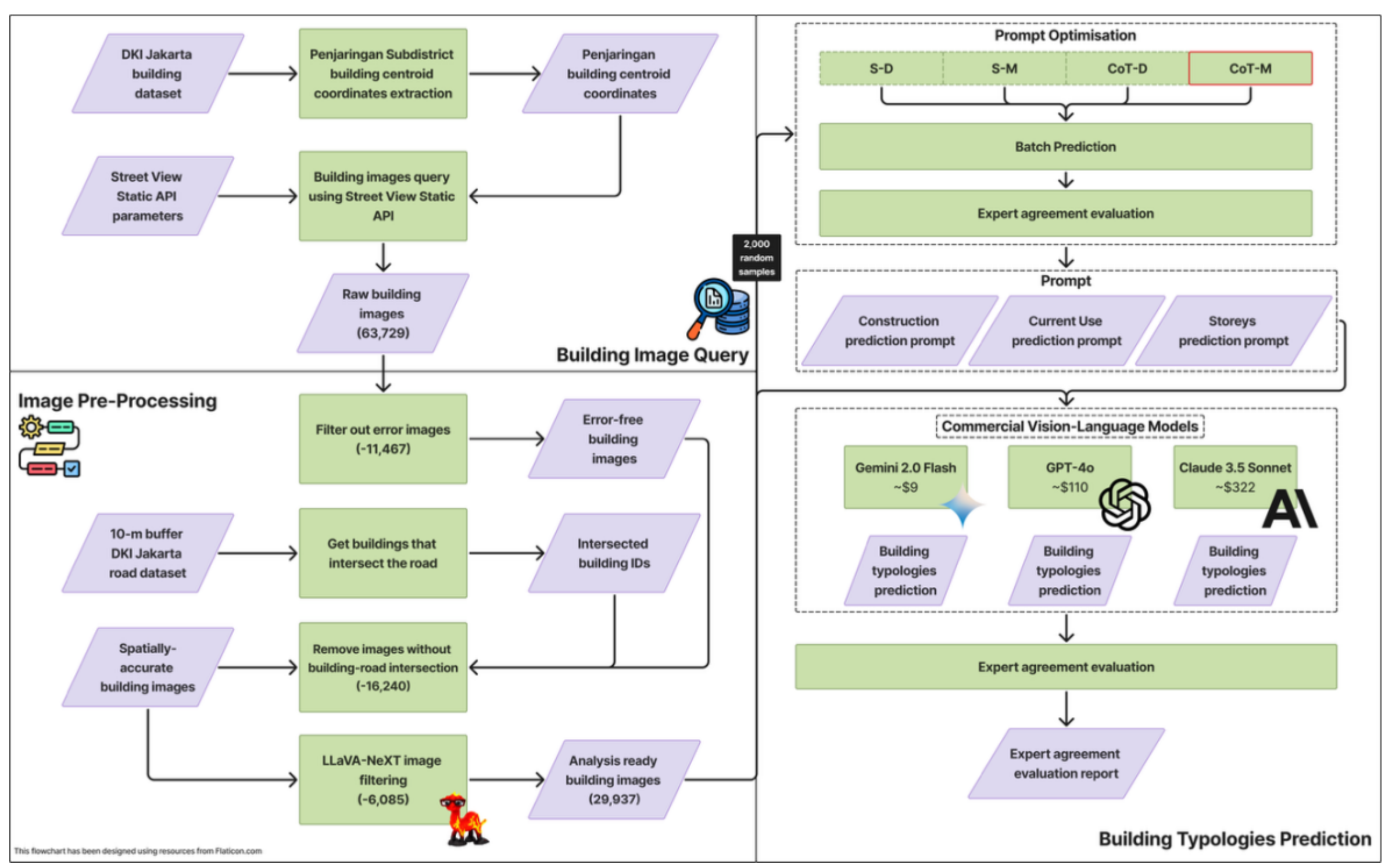

*Fig 2. Flowchart of the Methodology*

Initially, the VLM was tested on a small validation dataset of 2,000 images in a zero-shot setting to evaluate baseline reasoning and output quality, identifying potential limitations and areas for refinement as can be seen in Fig 3. These two thousand images were labelled manually by a team of experts (civil engineers and architects). Domain knowledge and familiarity with Indonesia's building typology were used to provide expert judgements. The labelling involved a quality assessment process by the research team, in which 20% of the sample was examined and labelled by consensus among experts to ensure the manual labelling was accurate. We ask each VLM model to provide a 200-word reasoning on why a specific prediction was made. The reasons were analysed by examining the probability of word occurrence, and 10 words with the highest probability were selected for each building typology prediction. This is then fed in as a precursor to check expert agreement with AI reasoning.

The analysis was then scaled to 6,000 images, in which expert agreement with model outputs was systematically assessed to validate agreement and consistency. Following this validation, the method was extended to the full dataset of 30,000 GSV images, enabling comprehensive spatial analysis across Penjaringan, North Jakarta. By examining where the VLM aligns with or diverges from expert reasoning, this approach provides deeper insights into how model outputs can be interpreted, trusted, and applied to study building features. Fig 3 shows the complete flow of the multi-stage framework scaling used in this paper.

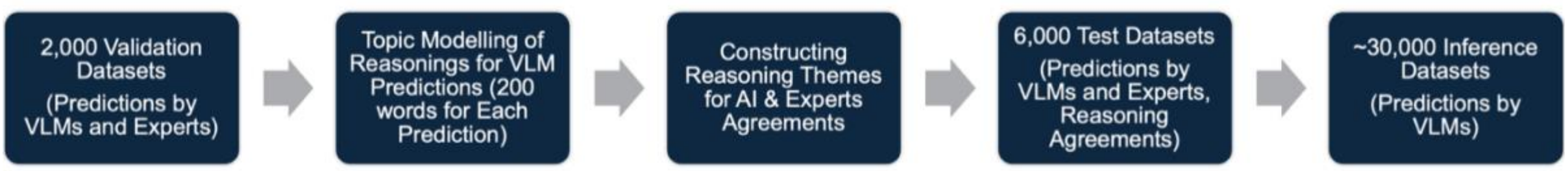

*Fig 3. Multi-stage Framework Scaling*

To optimise the performance of VLM models, a series of prompting strategies was employed to improve prompt structure, including zero-shot and Chain-of-Thought (CoT) prompting (Wei et al., 2022). This experiment also explores the implications of using both divided and merged prompts of the four building typologies predicted. The approaches for prompting are:

(1) Straight divided (s-d): running the prompts for each building typology separately using a zero-shot approach
(2) CoT divided (CoT-d): introducing CoT to encourage step-by-step thinking. Then, the prompts for each building typology are run separately using a zero-shot approach
(3) Straight merged (s-m): writing the prompts to predict each building typology in a merged prompt and running it for inference, and
(4) CoT merged (CoT-m): introducing CoT to encourage step-by-step thinking.

Then, the prompts to predict each building typology are combined into a single merged prompt and run for inference. Fig 4 and Table 1 provide a detailed description of the prompt strategies. The prompt samples are available in Fig A3 in the Appendix. The results of these three processes were evaluated through expert agreement evaluation and a confusion matrix.

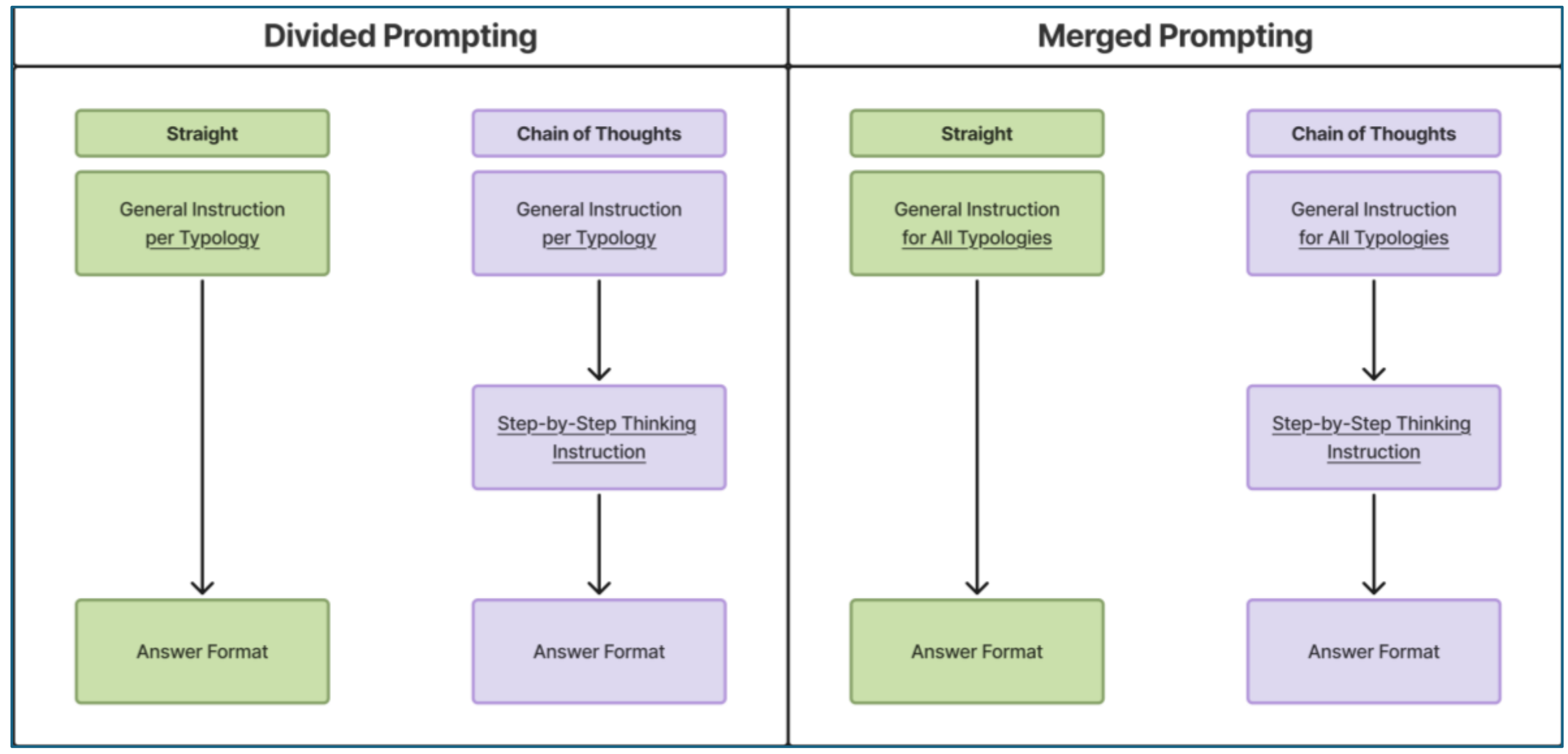

*Fig 4. Prompting Strategies Flow*

| Typologies to Predict | Prompting Technique | Prompt Block | Description |
|---|---|---|---|
| **Divided Prompting (Single-Task)** | Straight (s-d) | General Instruction | Abstraction of the task. |
| | | Answer Format | Instruction to answer in a predefined JSON format. |
| | Chain-of-Thought (CoT-d) | General Instruction | Abstraction of the task. |
| | | Step-by-Step Thinking Instruction | Instruction to think systematically by identifying building features, analysing surroundings, inferring a typology or typologies, and validating it/them against initial observations before finalising the prediction and reasoning. |
| | | Answer Format | Instruction to answer in a predefined JSON format. |
| **Merged Prompting (Multi-Task)** | Straight (s-m) | General Instruction | Abstraction of the task. |
| | | Answer Format | Instruction to answer in a predefined JSON format. |
| | Chain-of-Thought (CoT-m) | General Instruction | Abstraction of the task. |
| | | Step-by-Step Thinking Instruction | Instruction to think systematically by identifying building features, analysing surroundings, inferring a typology or typologies, and validating it/them against initial observations before finalising the prediction and reasoning. |
| | | Answer Format | Instruction to answer in a predefined JSON format. |

*Table 1. Prompting Strategies*

The prompt structure that yields the highest expert agreement is also used to ensure the quality of predictions for building typologies. The batch-processing principle is embedded in this process to increase inference efficiency and reduce costs by 50% per commercial VLM. The hyperparameters of each model (e.g., temperature, log probability, chat completion, and presence penalty) are set to their default values, except for the maximum tokens, which are adjusted to accommodate the selected prompt structure and to enable comprehensive reasoning in the prediction output. Temperature refers to a parameter that controls the randomness of the model's prediction, which can be adjusted to account for the variation of the predictions. The values might vary across models and versions, which we acknowledge as one of the study's limitations, which means that as a non-deterministic model, exact reproducibility might not be possible. The detailed API call parameters containing the temperature information are available in the accompanying code, and consistent results can still be achieved.

Each prediction generated by the visual-language models (VLMs) using the 2,000-validation dataset is accompanied by a 200-word textual rationale that outlines the model's underlying reasoning. To analyse these explanations, we examined the probability of words occurring in the reasons provided by VLM. We then selected 10 keywords based on their probability scores and frequencies in the VLMs' reasoning outputs. These keywords represent the most prominent concepts that influenced the models' predictions. To assess alignment with expert reasoning, we presented these keywords to domain experts and asked whether they accurately reflected the primary factors considered in their own decision-making processes. All experts have identified that the presented keywords were reasonable and can serve as a precursor to further analysis. The analysis expands to 6,000 test datasets, integrating predictions and reasoning alignments between VLMs and experts to assess the consistency and interpretability of predictive reasoning across cases. This comparison enabled us to determine the degree to which VLM-generated reasoning aligns with expert interpretative frameworks.

## 4. Results and Discussion

### 4.1. Zero-Shot Building Typology Predictions

Fig 5 presents the agreements between AI and human experts in predicting three building typologies: *Construction*, *Current Use, and Storeys*, as well as the cumulative *Overall Prediction* using the 2,000-validation dataset randomly selected from the overall dataset. Each heatmap quantifies the percentage of agreement (accuracy) between expert judgments and three VLMs under four prompt configurations: Straight Divided**,** CoT (Chain-of-Thought) Divided**,** Straight Merged**,** and CoT Merged**.** As described in the method section, we designed the study in different scales, and the purpose of the smaller-scale analysis is to create a baseline (1) to explore differences in prompt configuration and (2) to build a precursor containing the most frequent words used by VLM as reasoning for VLM and human expert agreements. In the first stage, both VLM and human experts were asked to analyse the building typologies.

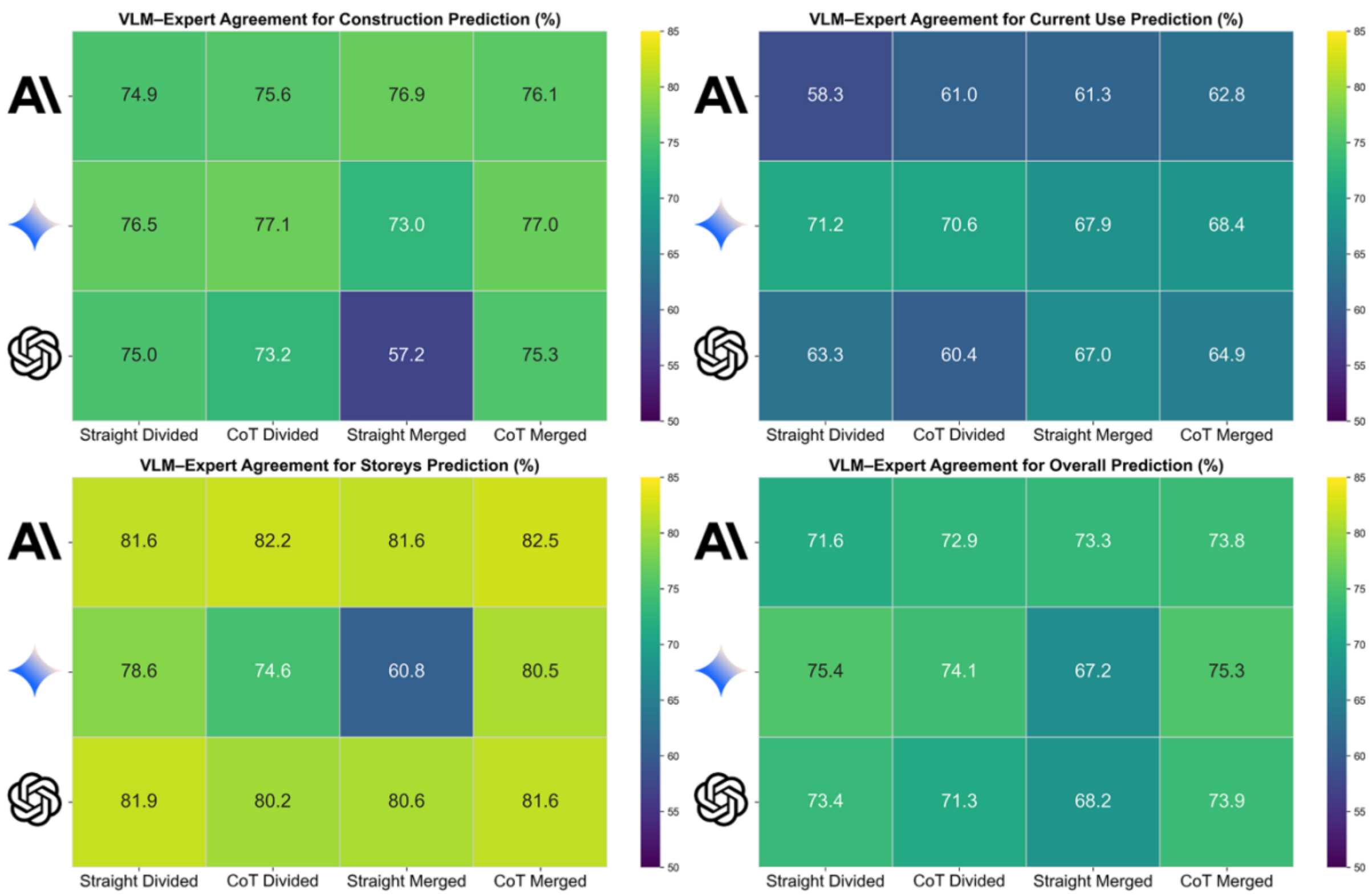

*Fig 5. VLM-expert Agreements (Representing Accuracy) with Different Prompt Strategies (n=2,000)*

Fig 5 highlights the agreements between the predictions made by VLM and those by human experts (architects and civil engineers). The colour gradients indicate the intensity of agreement, ranging from lower levels (blue/purple) to higher levels (yellow/green). The prediction for building storeys exhibits the highest overall agreement rates between VLMs and expert labelling, with several configurations surpassing 80% agreement, representing accuracy, which refers to the proportion of predictions that match the experts' labelling. In contrast, predictions for current building use show comparatively lower agreement, generally falling within the 60–70% range. This difference suggests that physical attributes, such as structural type and number of floors, are more easily inferred by AI from visual data, particularly when guided by step-by-step reasoning through straight divided and Chain-of-Thought (CoT) merged prompting. The confusion matrices for each prediction are shown in Appendix A1 and A2.

Our findings align with previous studies that have compared AI models with those of human experts investigating built environment analysis and urban studies. Using zero-shot learning, Belaroussi (2025) explores the capabilities of ChatGPT, Gemini, and Grok in visually assessing built environments and urban design and compares them with Architecture, Engineering, and Construction (AEC) experts. The study found that AI models tend to exhibit bias towards descriptive and functional attributes and require more nuanced training data to capture the diversity of human perceptions/predictions (Belaroussi, 2025). In this study, we find that lower performance on more complex tasks, such as assessing current-use prediction, likely stems from the inherent complexity and ambiguity in determining a building's function solely from its appearance. Unlike construction materials or storey count, which are visually explicit and objectively defined, current use often relies on subtle contextual cues such as signage, usage patterns, or interior layout that may not be clearly visible in images. Other studies, such as those by Ramalingam & Kumar (2025), highlighted the unique challenges of identifying current building usage in dense cities in the global South and proposed combining visual and textual data to improve prediction accuracy. Furthermore, categorising use types can be subjective, leading to greater variability in expert labelling and, subsequently, lower model alignment.

Table 2 shows the model's overall performance on our validation dataset. We use precision (how often positive results are correct), recall (how many true positives the model correctly identifies), and F1 (a score that balances precision and recall) as common metrics for evaluating a model. Our results show that the three VLMs perform similarly across prompt strategies, though there are some variations in their predictions. For example, the largest variation is in Straight Merged prompting for Construction Prediction (57.2%-76.9%) and Storeys Prediction (60.8%-81.6%). Moderate variation also occurs in Current Use prediction using Straight Divided prompting (58.3%-71.2%). The use of different prompts is exploratory rather than suggestive. Given the more stable performance when utilising CoT prompting, we decided to scale up the model only with the merged CoT prompts, as they are computationally less extensive than the divided prompts. It is worth noting that, in some cases, VLM models produced hallucinations, which are outputs that didn't align with the predefined answer categories for each typology. Throughout the experiments, we encountered hallucinations primarily when using Gemini with straight-divided prompts for the Construction (1.5% of predictions) and Current Use (1.2% of predictions) categories. These answers are flagged as "invalid" as shown in the confusion matrices in Fig A1.

| **Model** | | **Anthropic (Claude Sonnet)** | | | | **Google (Gemini)** | | | | **OpenAI (GPT)** | | | |
|---|---|---|---|---|---|---|---|---|---|---|---|---|---|
| | | **s-d** | **CoT-d** | **s-m** | **CoT-m** | **s-d** | **CoT-d** | **s-m** | **CoT-m** | **s-d** | **CoT-d** | **s-m** | **CoT-m** |
| Construction | Precision | 0.76 | 0.75 | 0.78 | 0.77 | 0.77 | 0.76 | 0.76 | 0.76 | 0.79 | 0.75 | 0.81 | 0.76 |
| | Recall | 0.75 | 0.76 | 0.77 | 0.76 | 0.76 | 0.77 | 0.73 | 0.77 | 0.75 | 0.73 | 0.57 | 0.75 |
| | F1 | 0.73 | 0.75 | 0.76 | 0.77 | 0.76 | 0.76 | 0.74 | 0.75 | 0.77 | 0.74 | 0.62 | 0.75 |
| Current Use | Precision | 0.78 | 0.74 | 0.76 | 0.75 | 0.78 | 0.78 | 0.77 | 0.76 | 0.78 | 0.75 | 0.76 | 0.75 |
| | Recall | 0.58 | 0.61 | 0.61 | 0.63 | 0.71 | 0.71 | 0.68 | 0.68 | 0.63 | 0.60 | 0.67 | 0.65 |

| | | | | | | | | | | | | |
|---|---|---|---|---|---|---|---|---|---|---|---|---|
| | F1 | 0.61 | 0.63 | 0.63 | 0.65 | 0.72 | 0.71 | 0.69 | 0.70 | 0.66 | 0.64 | 0.69 | 0.67 |
| Storeys | Precision | 0.82 | 0.82 | 0.82 | 0.82 | 0.82 | 0.81 | 0.82 | 0.82 | 0.83 | 0.84 | 0.85 | 0.85 |
| | Recall | 0.82 | 0.82 | 0.82 | 0.82 | 0.79 | 0.75 | 0.61 | 0.80 | 0.82 | 0.80 | 0.81 | 0.82 |
| | F1 | 0.79 | 0.81 | 0.8 | 0.81 | 0.79 | 0.77 | 0.69 | 0.81 | 0.8 | 0.82 | 0.81 | 0.82 |
| Overall | Precision | 0.79 | 0.77 | 0.78 | 0.78 | 0.79 | 0.79 | 0.78 | 0.79 | 0.80 | 0.78 | 0.81 | 0.79 |
| | Recall | 0.72 | 0.73 | 0.73 | 0.74 | 0.75 | 0.74 | 0.67 | 0.75 | 0.73 | 0.71 | 0.68 | 0.74 |
| | F1 | 0.71 | 0.73 | 0.73 | 0.74 | 0.76 | 0.75 | 0.71 | 0.75 | 0.74 | 0.73 | 0.71 | 0.75 |

*Table 2. Model Performance (n=2,000)*

Building on these results, we scale the framework to the 6,000-test dataset, focusing exclusively on predictions generated by VLMs. At this stage, the reasoning frameworks established in earlier phases are applied to guide the interpretation and validation of large-scale predictions. Overall, this stepwise process advances a comprehensive methodological pathway for evaluating, refining, and scaling reasoning-informed AI prediction systems. This method would allow us to understand how VLMs approach visual understanding and making judgements before concluding, and we can analyse what might drive poor prediction.

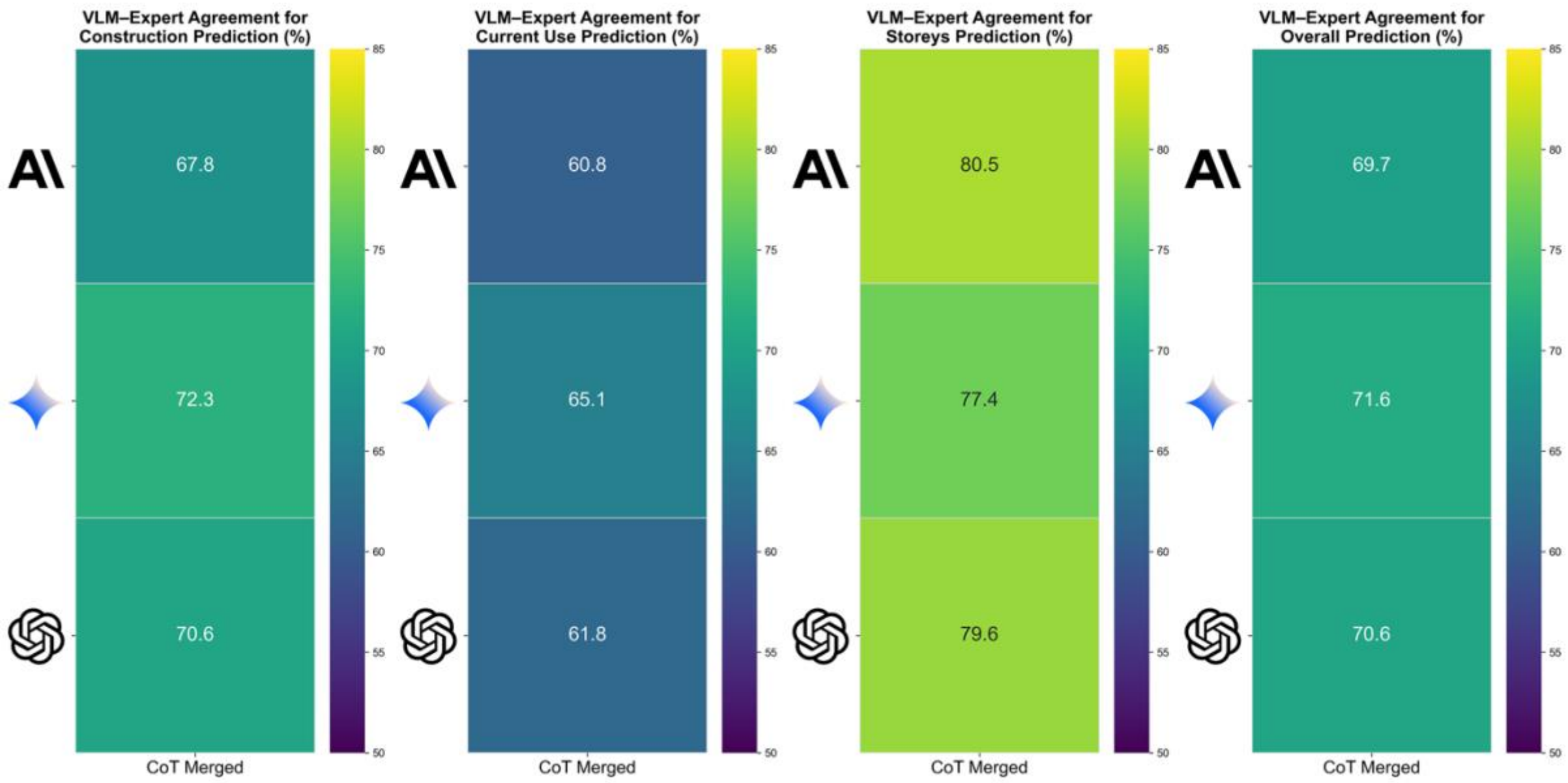

*Fig 6. VLM-Expert Agreement during Final Prediction (n=6,000)*

On the 6,000-image test set, Gemini 2.0 Flash achieved the highest expert agreement across all typologies (71.6%), followed closely by GPT-4o (70.6%) and Claude 3.5 Sonnet (69.7%), as shown in Fig 6. The small margin between Gemini and GPT-4o can be attributed to hallucination-induced disagreement in Gemini's outputs. Among the typologies, building storeys showed the highest overall expert agreement, followed by construction and current use. However, multiple instances of inaccurate predictions were observed, especially in identifying current building use. Mistakes were common when distinguishing between mixed-use and residential buildings, mainly because of their visual similarity in images, a challenge even for experts without on-site verification.

| Model | | Anthropic (Claude Sonnet) | Google (Gemini) | OpenAI (GPT) |
|---|---|---|---|---|
| Construction | Precision | 0.68 | 0.68 | 0.69 |
| | Recall | 0.68 | 0.72 | 0.71 |
| | F1 | 0.68 | 0.70 | 0.70 |
| Current Use | Precision | 0.68 | 0.67 | 0.68 |
| | Recall | 0.61 | 0.65 | 0.62 |
| | F1 | 0.62 | 0.65 | 0.63 |
| Storeys | Precision | 0.80 | 0.80 | 0.82 |
| | Recall | 0.81 | 0.77 | 0.80 |
| | F1 | 0.79 | 0.79 | 0.80 |
| Overall | Precision | 0.72 | 0.72 | 0.73 |
| | Recall | 0.70 | 0.72 | 0.71 |
| | F1 | 0.70 | 0.71 | 0.71 |

*Table 3. Model Performance (n=6,000)*

Table 3 shows the model's performance across the 6,000 test datasets. As the model is scaled up to include more images, the three models perform more similarly, with Gemini and GPT generally achieving higher F1 scores. GPT generally achieves higher precision, while Gemini is better at recall. All models consistently perform better at predicting building storeys.

### 4.2 Experts' Agreements in AI predictions

AI is often treated as a black box, and thus it is unable to explain how it reaches decisions or makes predictions (Fischer, 2022). Understanding the reasoning in Large Language Models (LLMs) has garnered attention (DeepSeek-AI et al., 2025; B. Wang et al., 2023), while in some applications, such as in health, LLMs often show correct predictions with wrong reasoning that do not align with experts' judgment (Maharana et al., 2025). This research aims to make the process more transparent and to explore the reasoning behind the VLM's predictions, examining whether the reasoning aligns with that of human experts. Fig 7 shows the word probabilities of the words used in the VLM reasoning for predicting building construction, current use, and storeys. It shows the different words used to arrive at the prediction, such as *material, climate, choice,* and *beam* for predicting construction materials; *entrance, design, neighbourhood* for predicting current building use; and *window, level, height, scale* for predicting building storeys. The initial word probability was presented and examined by experts to determine whether it is a reliable baseline for ground-truth checking, and the experts agreed that the words are reasonable.

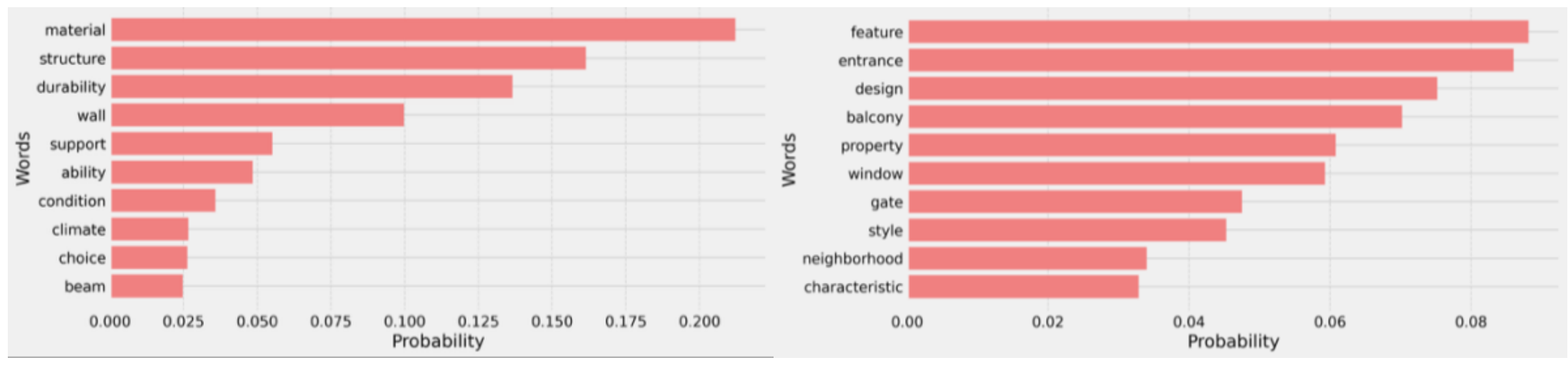

*(a) Construction* *(b) Current Use*

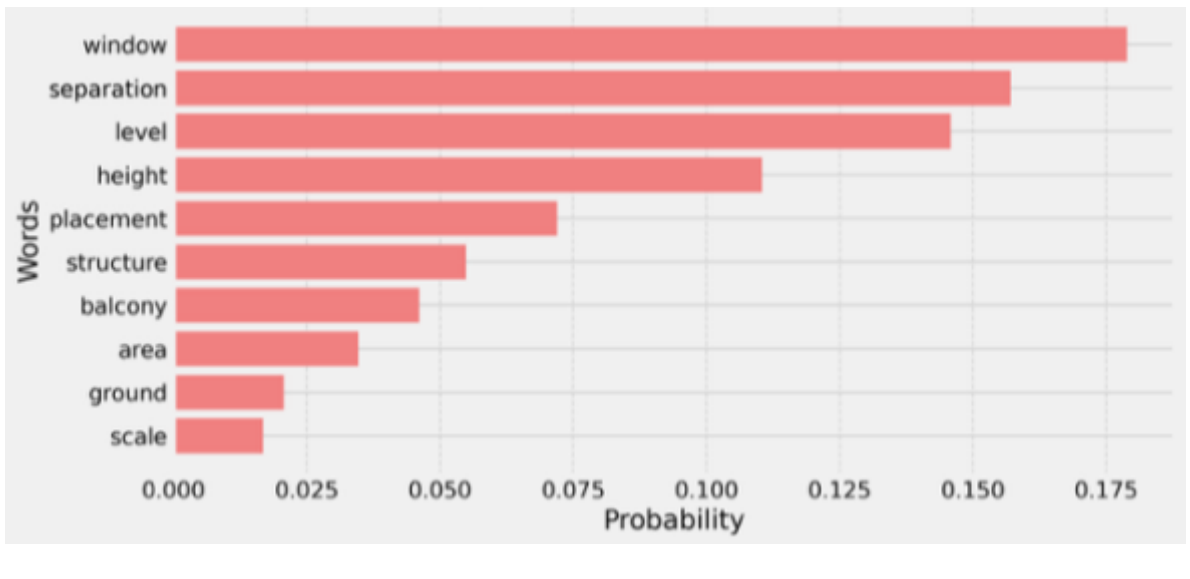

*(b) Storeys*

*Fig 7. Word Probability of VLM Reasonings (n=2000)*

In the next step of the analysis, we use the word probability as a precursor to how human experts create their building labelling. Fig 8 illustrates the differences between AI and human reasoning in predicting building typology, where the normalised frequencies highlight distinct reasoning tendencies. We notice that AI emphasises visually identifiable features and spatial indicators such as "*wall", "structure", "materials"* for the Construction category; "*entrance", "feature", "design", "balcony"* for the Current Use category. AI seems to place greater weight on tangible, measurable elements, reflecting its data-driven approach to recognising geometric and physical patterns. Human experts also rely on these visual cues, but there is a shift in the importance of a more contextual and localised understanding, such as *"condition", "support" and "choice"* for the Construction category and *"characteristic" and "neighbourhood"* for the Current Use category. Human experts place greater emphasis on interpretive aspects, suggesting awareness of design intent, material performance, and environmental adaptation. These differences are not too apparent in Storeys prediction, as storey counts are more identifiable than those in other categories, and thus rely on exterior elements such as *"window", "separation" and "height", which* are agreed by AI and human experts.

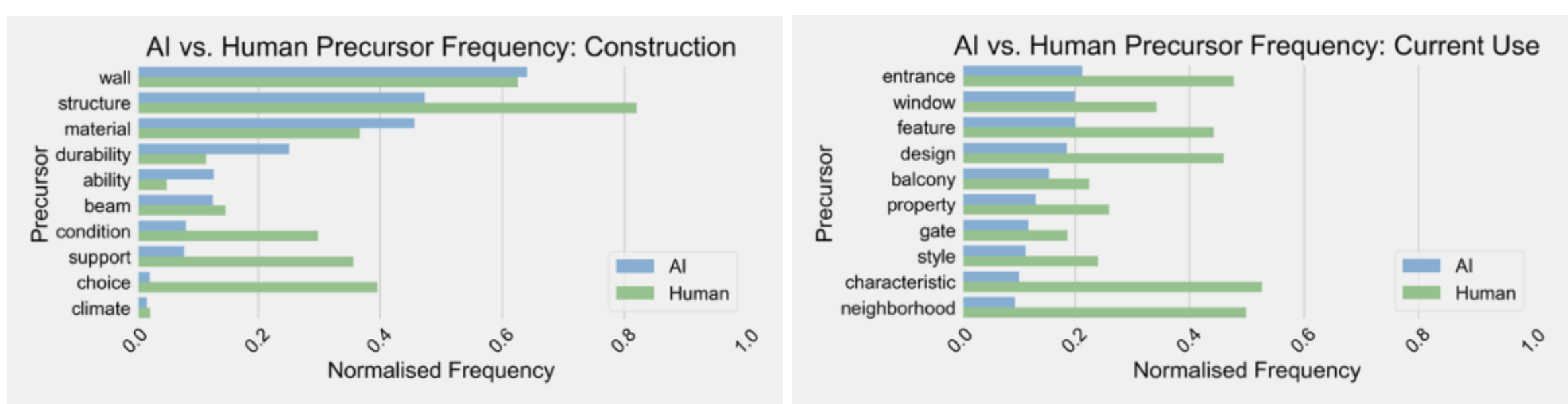

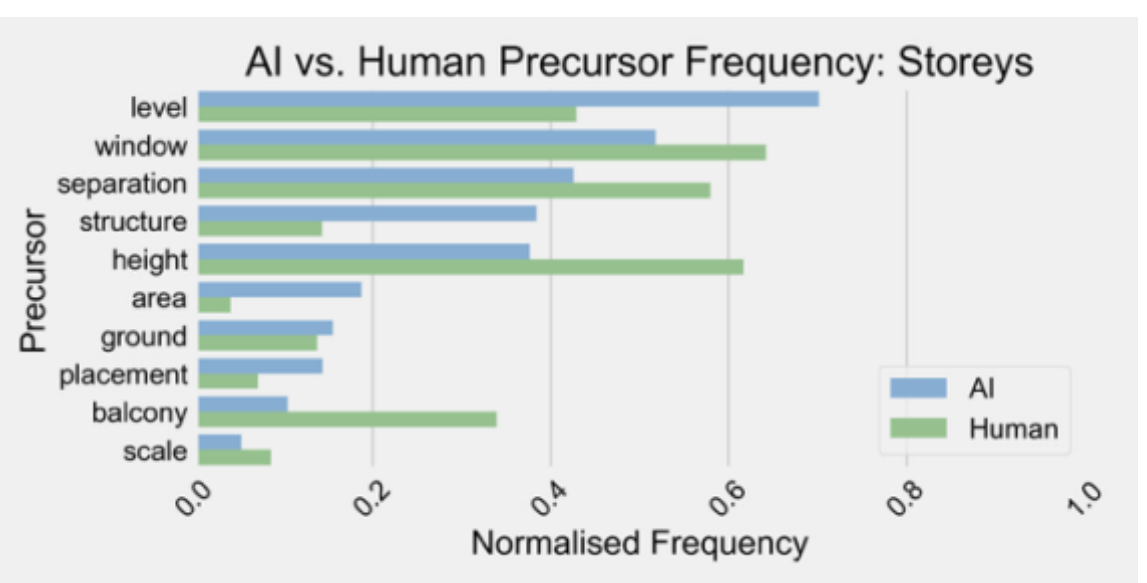

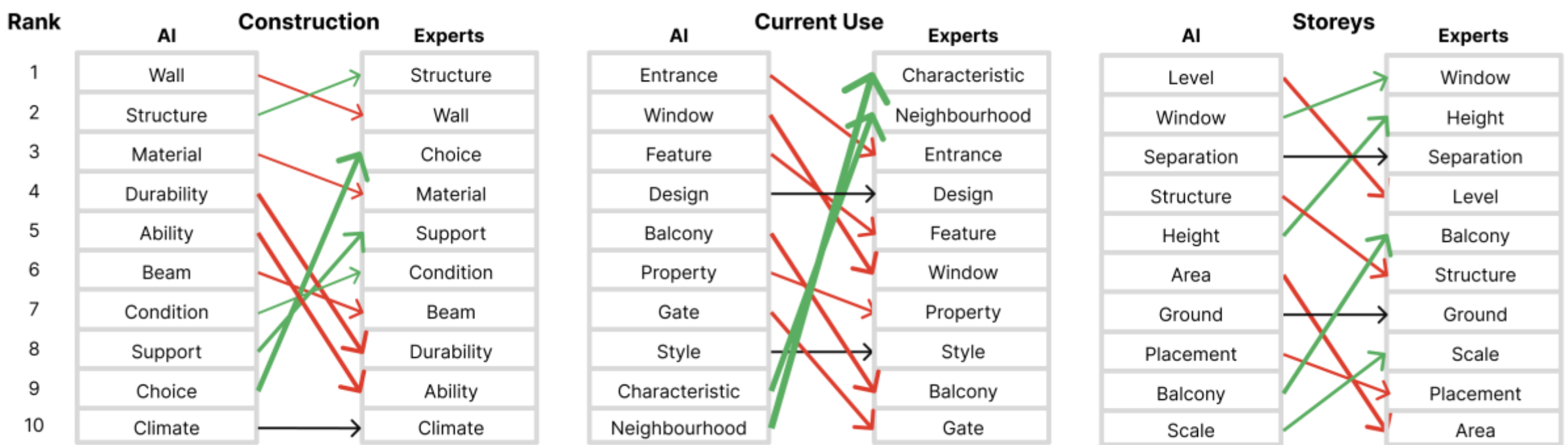

*Fig 8. AI vs Human Precursor (n=6,000)*

Fig 9 visualises the spatial distribution of predicted building typology in North Jakarta, Indonesia, using the full, cleaned dataset. Key areas of interest are magnified in inset panels, emphasising regions with diverse building compositions and visually communicating the area's typological variability. The map serves as a comprehensive overview of the built environment's typologies, potentially aiding urban planning, disaster resilience analysis, and infrastructure assessment. Using building typology prediction, we spatially map the urban fabric of North Jakarta with the assistance of AI. The smaller-scale datasets (n=2000 and n=6000) serve as benchmarks for AI prediction accuracy, enabling us to leverage AI to derive new information from visual street-view imagery in data-scarce environments, such as North Jakarta. This allows us to scale up detailed building information and provide reliable building-level information, although further verifications will be necessary to increase the accuracy level.

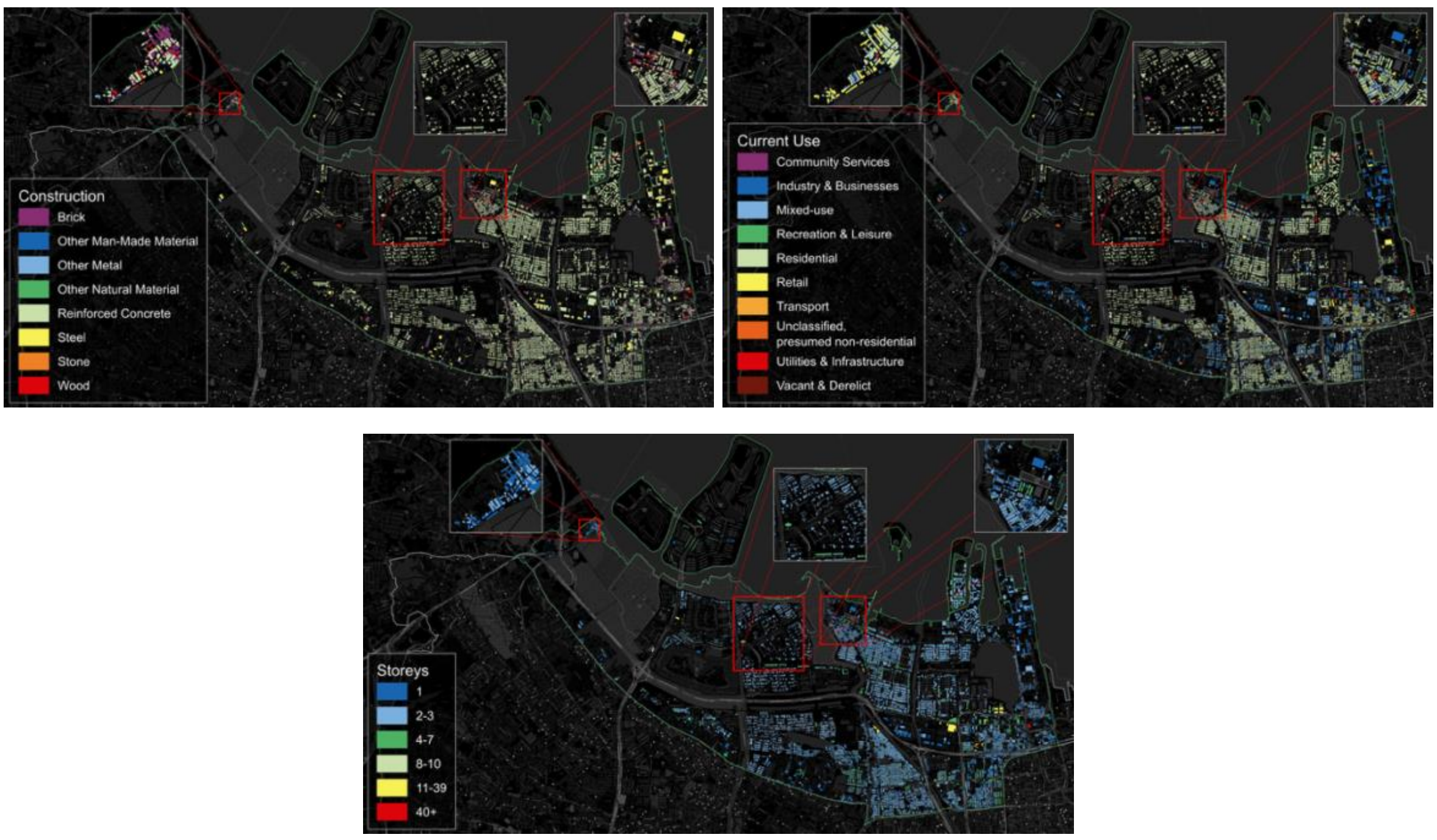

*Fig 9. Building Typology Inference Map - Construction (n=30,000)*

Fig 9 shows the spatial distribution of buildings by construction material, current use, and number of storeys. Based on our analysis, buildings in North Jakarta are mainly constructed with reinforced concrete and bricks. Some denser, more informal coastal settlement areas, such as Kamal Muara and Muara Angke, highlighted on the maps, use a more diverse range of building materials, including some steel, stone, and wood, as many buildings are built organically by the communities. These are areas with longstanding critical environmental issues due to flooding and land subsidence, where some traditional housing, such as wooden stilt housing, remains evident, supporting maritime-oriented livelihoods (Colven, 2020a; Husin, 2025). Understanding these spatial configurations could also help quantify risks, such as those from frequent coastal flooding. In terms of building storeys, North Jakarta is dominated by low-rise buildings between 1 and 3 storeys, with residential use followed by industry and businesses dominating the current use of the buildings. Various research has indicated that construction materials, building heights or the number of storeys, and how buildings are utilised are important factors in determining building vulnerabilities to hazard exposure, such as flood, but such data are not always available (Balasbaneh et al., 2020; Englhardt et al., 2019; Godfrey et al., 2015). This is particularly important in the Global South context, where building stocks are diverse and often irregular (Englhardt et al., 2019). Overall, this research provides a basis for further studies that require detailed building-level datasets and contributes to understanding the opportunities and challenges of AI automation for building typology prediction.

## 5. Conclusions and Further Study

This study contributes to the growing field of automated urban analysis by examining the capabilities of Vision-Language Models (VLMs) for predicting building typology using building-level visual data. By comparing several VLM approaches to human expert judgments, we provide a foundational understanding of how these models interpret architectural features and how effectively they can replicate or augment human decision-making in urban analytics. Our findings suggest that scalable, AI-driven approaches to typology classification have the potential to enhance the efficiency and consistency of data creation for urban modelling and spatial planning, although further validation techniques should be implemented to improve the overall accuracy. This is particularly valuable in contexts where manual annotation is time-consuming, costly, or infeasible at scale. The potential for integrating VLMs into geospatial pipelines points to a future of more automated, data-rich urban environments better suited to responsive policymaking and planning, such as in building typology prediction to understand detailed urban fabrics.

However, one of the primary challenges identified in this study is the reliance on visual features as the sole or primary input for classification. While they provide helpful cues, they may not fully capture the function or use of buildings, especially in mixed-use or visually ambiguous cases. This limitation highlights the need for multi-modal approaches. For instance, the work by Ramalingam & Kumar (2025) has demonstrated the effectiveness of combining vision-based inputs with textual data extracted from building signage, advertisements, and banners. Their meta-classifier approach leverages this additional context to enhance classification agreements and interpretability, indicating a promising direction for future research.

In conclusion, the integration of VLMs into urban modelling workflows holds significant promise. Still, it must be approached with attention to the complexities of architectural form, urban context, and data heterogeneity. Moving forward, interdisciplinary collaboration between computer scientists, urban planners, and domain experts will be essential to refine these models, improve their interpretability, and ensure their responsible application in real-world settings. Continued exploration of multi-modal learning, transfer learning, and cross-domain adaptation will be crucial for advancing the state of the art in automated urban analytics.

## References

Allen, G. C. (2020). Understanding AI Technology. *Joint Artificial Intelligence Center (JAIC)*, (April).

Ananthram, A., Stengel-Eskin, E., Bansal, M., & McKeown, K. (2025). *See It from My Perspective: How Language Affects Cultural Bias in Image Understanding*. http://arxiv.org/abs/2406.11665

Anguelov, D., Dulong, C., Filip, D., Frueh, C., Lafon, S., Lyon, R., Ogale, A., Vincent, L., & Weaver, J. (2010). Google street view: Capturing the world at street level. *Computer*, *43*(6). https://doi.org/10.1109/MC.2010.170

Barredo Arrieta, A., Tabik, S., García López, S., Molina Cabrera, D., Herrera Triguero, F., & Díaz Rodríguez, N. A. (2019). *Explainable Artificial Intelligence (XAI): Concepts, taxonomies, opportunities and challenges toward responsible AI*. https://digibug.ugr.es/handle/10481/77982

Biljecki, F., & Ito, K. (2021). Street view imagery in urban analytics and GIS: A review. In *Landscape and Urban Planning* (Vol. 215). https://doi.org/10.1016/j.landurbplan.2021.104217

Blagojević, N., Brzev, S., Petrović, M., Borozan, J., Bulajić, B., Marinković, M., Hadzima-Nyarko, M., Koković, V., & Stojadinović, B. (2023). Residential building stock in Serbia: classification and vulnerability for seismic risk studies. *Bulletin of Earthquake Engineering*, *21*(9). https://doi.org/10.1007/s10518-023-01676-0

Brown, T. B., Mann, B., Ryder, N., Subbiah, M., Kaplan, J., Dhariwal, P., Neelakantan, A., Shyam, P., Sastry, G., Askell, A., Agarwal, S., Herbert-Voss, A., Krueger, G., Henighan, T., Child, R., Ramesh, A., Ziegler, D. M., Wu, J., Winter, C., … Amodei, D. (2020). Language models are few-shot learners. *Advances in Neural Information Processing Systems*, *2020-December*.

Cerezo, C., Sokol, J., AlKhaled, S., Reinhart, C., Al-Mumin, A., & Hajiah, A. (2017). Comparison of four building archetype characterization methods in urban building energy modeling (UBEM): A residential case study in Kuwait City. *Energy and Buildings*, *154*. https://doi.org/10.1016/j.enbuild.2017.08.029

Colven, E. (2020). Thinking beyond success and failure: Dutch water expertise and friction in postcolonial Jakarta. *Environment and Planning C: Politics and Space*, *38*(6). https://doi.org/10.1177/2399654420911947

Fan, Z., Feng, C. C., & Biljecki, F. (2025). Coverage and bias of street view imagery in mapping the urban environment. *Computers, Environment and Urban Systems*, *117*, 102253. https://doi.org/10.1016/J.COMPENVURBSYS.2025.102253

Firmansyah, H. R., Sarli, P. W., Twinanda, A. P., Santoso, D., & Imran, I. (2024). Building typology classification using convolutional neural networks utilizing multiple ground-level image process for city-scale rapid seismic vulnerability assessment. *Engineering Applications of Artificial Intelligence*, *131*. https://doi.org/10.1016/j.engappai.2023.107824

Fischer, G. (2022). A Research Framework Focused on AI and Humans instead of AI versus Humans. *CEUR Workshop Proceedings*, *3136*.

Goel, R., Garcia, L. M. T., Goodman, A., Johnson, R., Aldred, R., Murugesan, M., Brage, S., Bhalla, K., & Woodcock, J. (2018). Estimating city-level travel patterns using street imagery: A case study of using Google Street View in Britain. *PLoS ONE*, *13*(5). https://doi.org/10.1371/journal.pone.0196521

Gonzalez, D., Rueda-Plata, D., Acevedo, A. B., Duque, J. C., Ramos-Pollán, R., Betancourt, A., & García, S. (2020). Automatic detection of building typology using deep learning methods on street level images. *Building and Environment*, *177*. https://doi.org/10.1016/j.buildenv.2020.106805

Hecht, R., Meinel, G., & Buchroithner, M. (2015). Automatic identification of building types based on topographic databases–a comparison of different data sources. *International Journal of Cartography*, *1*(1). https://doi.org/10.1080/23729333.2015.1055644

Hoffmann, E. J., Wang, Y., Werner, M., Kang, J., & Zhu, X. X. (2019). Model fusion for building type classification from aerial and street view images. *Remote Sensing*, *11*(11). https://doi.org/10.3390/rs11111259

Huang, T., Liu, Z., Wang, R., Zhang, Y., & Jing, L. (2025). Visual hallucination detection in large vision-language models via evidential conflict. *International Journal of Approximate Reasoning*, *186*, 109507. https://doi.org/10.1016/J.IJAR.2025.109507

Kaduri, O., Bagon, S., & Dekel, T. (2024). What's in the Image? A Deep-Dive into the Vision of Vision Language Models. *ArXiv*.

Karunaratne, G. (2023). Resilient Urbanism. *Bhumi, The Planning Research Journal*, *10*(1). https://doi.org/10.4038/bhumi.v10i1.101

Keralis, J. M., Javanmardi, M., Khanna, S., Dwivedi, P., Huang, D., Tasdizen, T., & Nguyen, Q. C. (2020). Health and the built environment in United States cities: Measuring associations using Google Street View-derived indicators of the built environment. *BMC Public Health*, *20*(1). https://doi.org/10.1186/s12889-020-8300-1

Kumar, Y., Koul, A., Singla, R., & Ijaz, M. F. (2023). Artificial intelligence in disease diagnosis: a systematic literature review, synthesizing framework and future research agenda. *Journal of Ambient Intelligence and Humanized Computing*, *14*(7). https://doi.org/10.1007/s12652-021-03612-z

Lang, D. H., Kumar, A., Sulaymanov, S., & Meslem, A. (2018). Building typology classification and earthquake vulnerability scale of Central and South Asian building stock. *Journal of Building Engineering*, *15*. https://doi.org/10.1016/j.jobe.2017.11.022

Lartey, D., & Law, K. M. Y. (2025). Artificial intelligence adoption in urban planning governance: A systematic review of advancements in decision-making, and policy making. *Landscape and Urban Planning*, *258*, 105337. https://doi.org/10.1016/J.LANDURBPLAN.2025.105337

Law, S., Paige, B., & Russell, C. (2019). Take a look around: Using street view and satellite images to estimate house prices. *ACM Transactions on Intelligent Systems and Technology*, *10*(5). https://doi.org/10.1145/3342240

Liu, S., Cheng, H., Liu, H., Zhang, H., Li, F., Ren, T., Zou, X., Yang, J., Su, H., Zhu, J., Zhang, L., Gao, J., & Li, C. (2023). *LLaVA-Plus: Learning to Use Tools for Creating Multimodal Agents*. http://arxiv.org/abs/2311.05437

Nguyen, Q. C., Keralis, J. M., Dwivedi, P., Ng, A. E., Javanmardi, M., Khanna, S., Huang, Y., Brunisholz, K. D., Kumar, A., & Tasdizen, T. (2021). Leveraging 31 Million Google Street View Images to Characterize Built Environments and Examine County Health Outcomes. *Public Health Reports*, *136*(2). https://doi.org/10.1177/0033354920968799

Nishida, N., Yamakawa, M., Shiina, T., Mekada, Y., Nishida, M., Sakamoto, N., Nishimura, T., Iijima, H., Hirai, T., Takahashi, K., Sato, M., Tateishi, R., Ogawa, M., Mori, H., Kitano, M., Toyoda, H., Ogawa, C., & Kudo, M. (2022). Artificial intelligence (AI) models for the ultrasonographic diagnosis of liver tumors and comparison of diagnostic accuracies between AI and human experts. *Journal of Gastroenterology*, *57*(4). https://doi.org/10.1007/s00535-022-01849-9

Nofal, O., Rosenheim, N., Kameshwar, S., Patil, J., Zhou, X., van de Lindt, J. W., Duenas-Osorio, L., Cha, E. J., Endrami, A., Sutley, E., Cutler, H., Lu, T., Wang, C., & Jeon, H. (2024). Community-level post-hazard functionality methodology for buildings exposed to floods. *Computer-Aided Civil and Infrastructure Engineering*, *39*(8). https://doi.org/10.1111/mice.13135

Perez, J., Fusco, G., Araldi, A., & Fuse, T. (2018). Building Typologies for Urban Fabric Classification : Osaka and Marseille Case Studies. *International Conference on Spatial Analysis and Modeling*.

Radford, A., Kim, J. W., Hallacy, C., Ramesh, A., Goh, G., Agarwal, S., Sastry, G., Askell, A., Mishkin, P., Clark, J., Krueger, G., & Sutskever, I. (2021). Learning Transferable Visual Models From Natural Language Supervision. *Proceedings of Machine Learning Research*, *139*.

Ramalingam, S. P., & Kumar, V. (2025). Building usage prediction in complex urban scenes by fusing text and facade features from street view images using deep learning. *Building and Environment*, *267*, 112174. https://doi.org/10.1016/J.BUILDENV.2024.112174

Reinhart, C. F., & Cerezo Davila, C. (2016). Urban building energy modeling - A review of a nascent field. In *Building and Environment* (Vol. 97). https://doi.org/10.1016/j.buildenv.2015.12.001

Rzotkiewicz, A., Pearson, A. L., Dougherty, B. V., Shortridge, A., & Wilson, N. (2018). Systematic review of the use of Google Street View in health research: Major themes, strengths, weaknesses and possibilities for future research. In *Health and Place* (Vol. 52). https://doi.org/10.1016/j.healthplace.2018.07.001

S, H., W, Z., H, L., P, H., & P, J. (2019). Spatial Predictions of Active Travel Based on Google Street View Imagery and Place of Interest Data. *Environmental Epidemiology*, *3*(Supplement 1). https://doi.org/10.1097/01.ee9.0000607440.14860.a9

Sakti, A. D., Ihsan, K. T. N., Anggraini, T. S., Shabrina, Z., Sasongko, N. A., Fachrizal, R., Aziz, M., Aryal, J., Yuliarto, B., Hadi, P. O., & Wikantika, K. (2022). Multi-Criteria Assessment for City-Wide Rooftop Solar PV Deployment: A Case Study of Bandung, Indonesia. *Remote Sensing*, *14*(12). https://doi.org/10.3390/rs14122796

Sun, M., Zhang, F., & Duarte, F. (2021). Automatic Building Age Prediction from Street View Images. *Proceedings of 2021 7th IEEE International Conference on Network Intelligence and Digital Content, IC-NIDC 2021*. https://doi.org/10.1109/IC-NIDC54101.2021.9660554

Turpin, M., Michael, J., Perez, E., & Bowman, S. R. (2023). Language Models Don't Always Say What They Think: Unfaithful Explanations in Chain-of-Thought Prompting. *Advances in Neural Information Processing Systems.*, *36*, 74952–74965.

Umar, F., Amoah, J., Asamoah, M., Dzodzomenyo, M., Igwenagu, C., Okotto, L. G., Okotto-Okotto, J., Shaw, P., & Wright, J. (2023). On the potential of Google Street View for environmental waste quantification in urban Africa: An assessment of bias in spatial coverage. *Sustainable Environment*, *9*(1). https://doi.org/10.1080/27658511.2023.2251799

Wang, G. Y. (2023). The effect of environment on housing prices: Evidence from the Google Street View. *Journal of Forecasting*, *42*(2). https://doi.org/10.1002/for.2907

Wei, J., Wang, X., Schuurmans, D., Bosma, M., Ichter, B., Xia, F., Chi, E. H., Le, Q. V., & Zhou, D. (2022). Chain-of-Thought Prompting Elicits Reasoning in Large Language Models. *Advances in Neural Information Processing Systems*, *35*.

Xiao, J., Wu, W., Zhao, J., Fang, M., & Wang, J. (2025). Enhancing long-form question answering via reflection with question decomposition. *Information Processing & Management*, *62*(6), 104274. https://doi.org/10.1016/J.IPM.2025.104274

Yin, L., & Wang, Z. (2016). Measuring visual enclosure for street walkability: Using machine learning algorithms and Google Street View imagery. *Applied Geography*, *76*. https://doi.org/10.1016/j.apgeog.2016.09.024

Yin, Z., & Wang, S. (2025). Enhancing scientific table understanding with type-guided chain-of-thought. *Information Processing & Management*, *62*(4), 104159. https://doi.org/10.1016/J.IPM.2025.104159

Zhang, Y., & Dong, R. (2018). Impacts of street-visible greenery on housing prices: Evidence from a hedonic price model and a massive street view image dataset in Beijing. *ISPRS International Journal of Geo-Information*, *7*(3). https://doi.org/10.3390/ijgi7030104

**Codes Availability**

**https://github.com/zarashabrina/AI-Experts-Agreements**

**Appendix**

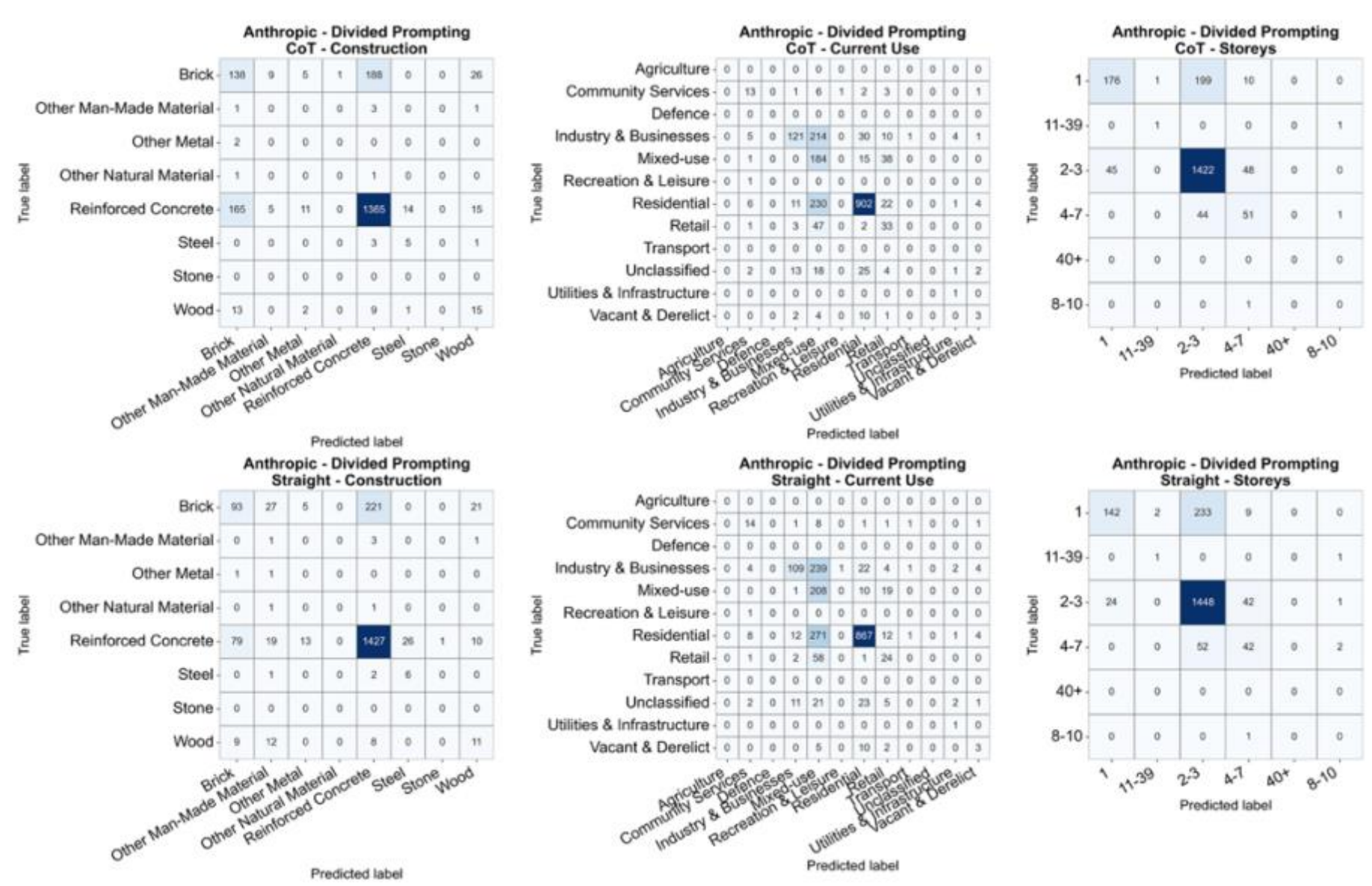

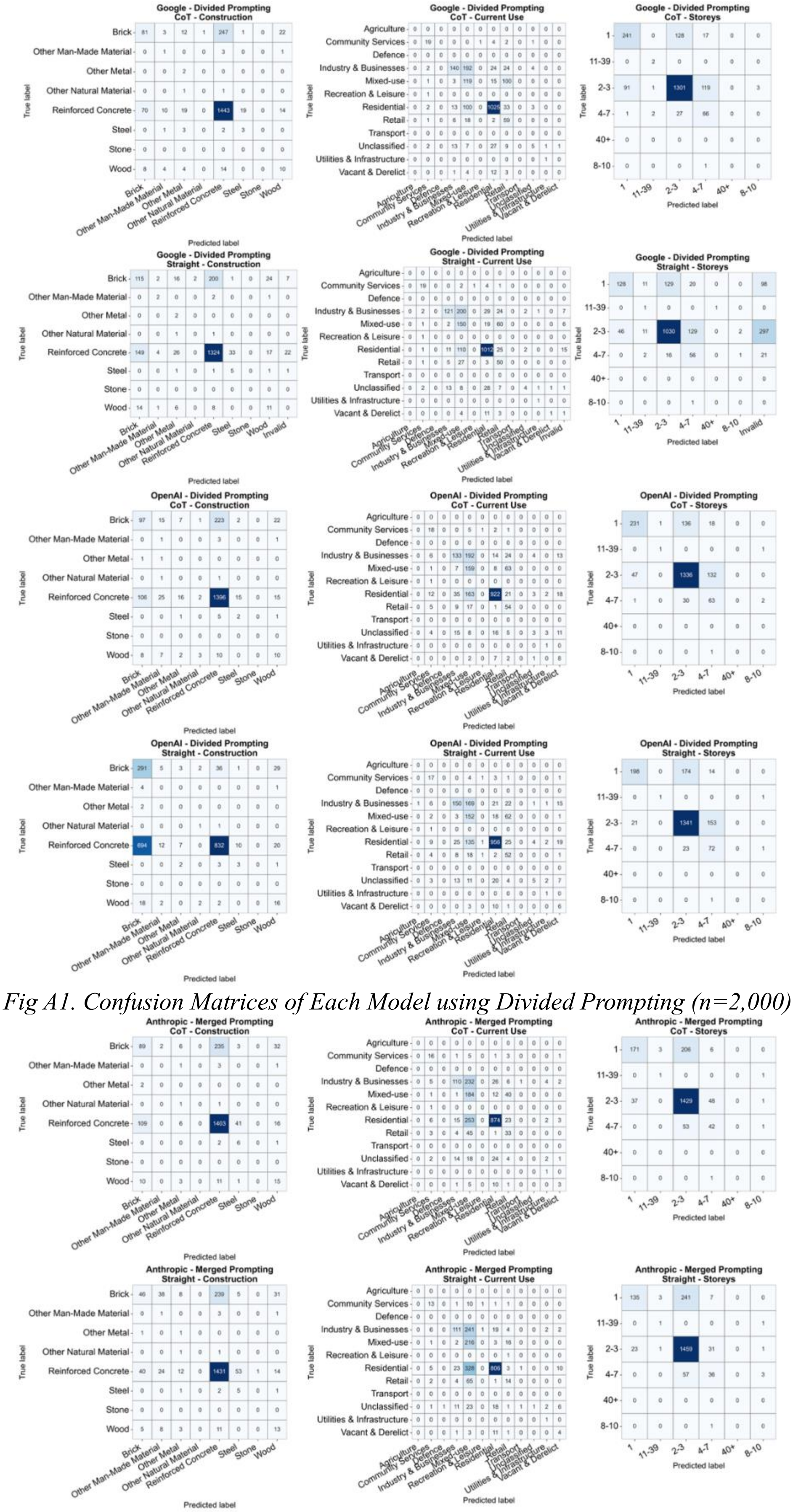

*Fig A1. Confusion Matrices of Each Model using Divided Prompting (n=2,000)*

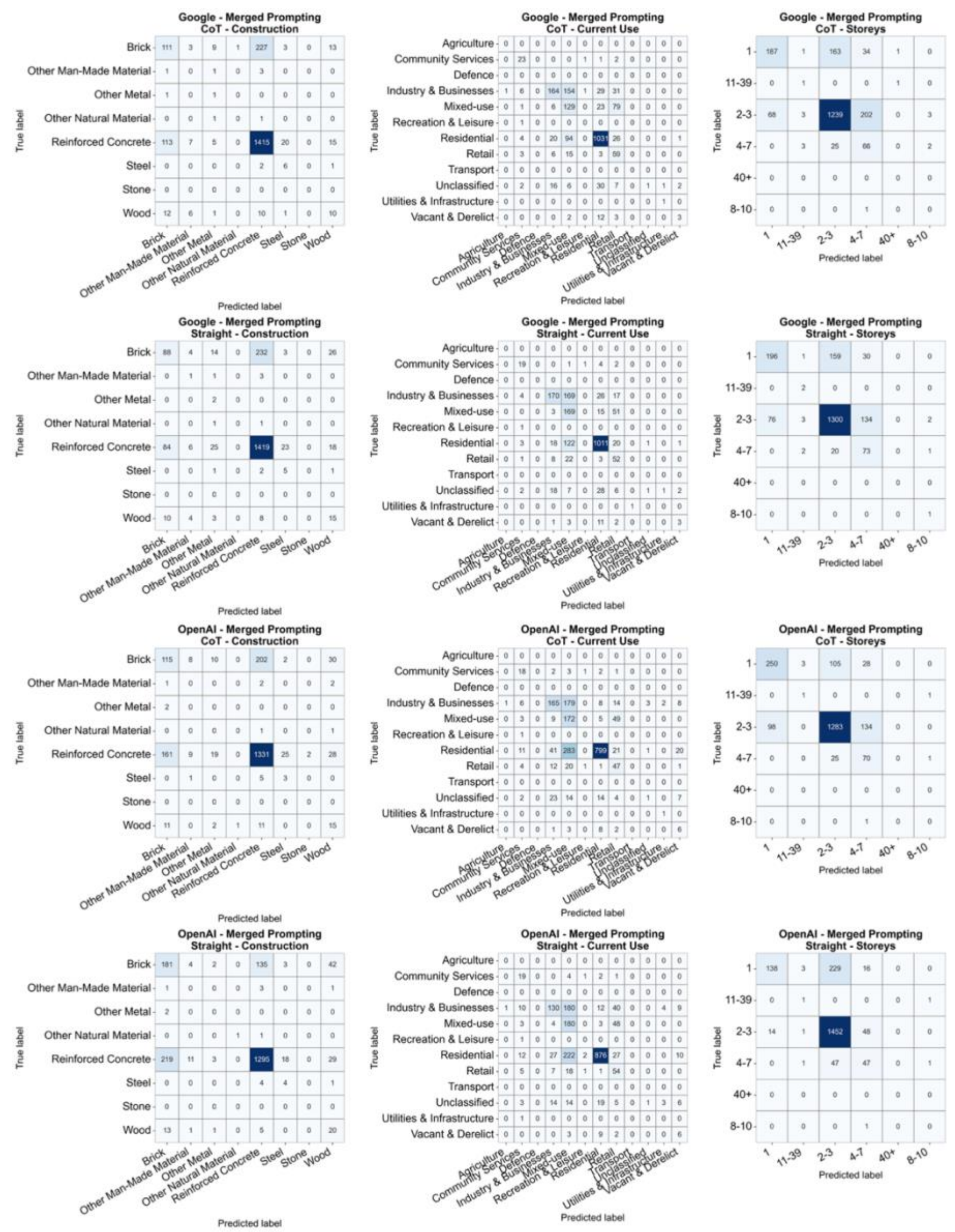

*Fig A2. Confusion Matrices of Each Model using Merged Prompting (n=2,000)*

```
<instructions>
You will be presented with a building image. Please infer the most likely
<building_construction>.
</instructions>

<formatting>
Only select <building_construction> from this list: ["Wood", "Stone", "Brick",
"Steel", "Reinforced Concrete", "Other Metal", "Other Natural Material", "Other
Man-Made Material"].

Organize your answer in the following JSON format containing 2 keys:
{
    "construction": <building_construction>,
    "reason": <reasoning>
}

The meaning of 2 keys:
- "construction": the most likely <building_construction> chosen from the provided
list.
- "reason": a concise explanation supporting your prediction, considering visible
material characteristics, structural features, and any other relevant details
specific to buildings located in North Jakarta, Indonesia (please do not use line
breaks in the reason and answer in 400 characters or 75 tokens max).
</formatting>
```

*(a) Sample Prompts for Straight Divided*

```
<instructions>
Your task is to predict the construction type of a building based on the image provided by users.

Instructions:
1. You will be presented with <building>, an image containing a main building located roughly at the
center of the image.
2. The image quality should be clear enough to identify building materials and structural features
specific to buildings located in North Jakarta, Indonesia.
3. Your goal is to infer the most likely <building_construction>.
4. If it's not possible to infer the most likely <building_construction> from the main building in
the centre of the image, infer from surrounding buildings with similar characteristics or features.
</instructions>

<step_by_step_thinking>
Think before you predict the building construction in <thinking>.

Structured thinking guide:
1. Identify and describe features of the building relevant to the building construction prediction.
2. Analyse the building's surroundings to add context for the building construction prediction.
3. Infer the most suitable building construction.
4. Compare and validate the predicted building construction with the features and surrounding
environment you described earlier (correct the previous prediction, if necessary).

After comparison and validation, proceed to infer <building_construction> and the <reasoning> behind
it.
</step_by_step_thinking>

<formatting>
Only select <building_construction> from this list: ["Wood", "Stone", "Brick", "Steel", "Reinforced
Concrete", "Other Metal", "Other Natural Material", "Other Man-Made Material"].

Organize your answer in the following JSON format containing 3 keys:
{
    "thoughts": <thinking>,
    "construction": <building_construction>,
    "reason": <reasoning>
}

The meaning of 3 keys:
- "thoughts": detailed, step-by-step breakdown of your reasoning process following the 4 points in
the structured thinking guide (each thinking step should be at around 400 characters or 75 tokens
max).
- "construction": the most likely <building_construction> chosen from the provided list.
- "reason": a concise explanation supporting your prediction, considering visible material
characteristics, structural features, and any other relevant details specific to buildings located
in North Jakarta, Indonesia (please do not use line breaks in the reason and answer in 400
characters or 75 tokens max).
</formatting>
```

*(b) Sample Prompts for Chain-of-Thought Divided*

```
<instructions>
You will be presented with a building image. Please infer the most likely <building_construction>,
<building_current_use>, and <building_storeys>.
</instructions>

<formatting>
Only select <building_construction> from this list: ["Wood", "Stone", "Brick", "Steel", "Reinforced
Concrete", "Other Metal", "Other Natural Material", "Other Man-Made Material"].
Only select <building_current_use> from this list: ["Mixed-use", "Residential", "Retail", "Industry
& Businesses", "Community Services", "Recreation & Leisure", "Transport", "Utilities &
Infrastructure", "Defence", "Agriculture", "Vacant & Derelict", "Unclassified, presumed
non-residential"].
Only select <building_storeys> from this list: ["1", "2-3", "4-7", "8-10", "11-39", "40+"].

Organize your answer in the following JSON format containing 6 keys:
{
    "construction": <building_construction>,
    "reason_construction": <reasoning_construction>,
    "currentuse": <building_current_use>,
    "reason_currentuse": <reasoning_current_use>,
    "storeys": <building_storeys>,
    "reason_storeys": <reasoning_storeys>
}

The meaning of 6 keys:
"construction": the most likely <building_construction> chosen from the provided list.
"reason_construction": a concise explanation supporting your prediction, considering visible
material characteristics, structural features, and any other relevant details specific to buildings
located in North Jakarta, Indonesia (please do not use line breaks in the reason and answer in 400
characters or 75 tokens max).
"currentuse": the most likely <building_current_use> chosen from the provided list.
"reason_currentuse": a concise explanation supporting your prediction, considering visible features,
usage indications, signage, layout, and any other relevant details specific to buildings located in
North Jakarta, Indonesia (please do not use line breaks in the reason and answer in 400 characters
or 75 tokens max).
"storeys": the most likely <building_storeys> chosen from the provided list.
"reason_storeys": a concise explanation supporting your prediction, considering visible storeys,
structural elements like windows and floor separations, and any other relevant details specific to
buildings located in North Jakarta, Indonesia (please do not use line breaks in the reason and
answer in 400 characters or 75 tokens max).
</formatting>
```

*(c) Sample Prompts for Straight Merged*

```
<instructions>
Your task is to predict building typologies based on the image provided by users.

Instructions:
1. You will be presented with <building>, an image containing a main building located roughly at the center of the
image.
2. The image quality should be clear enough to identify 3 different building typologies
2.1. Construction: building materials and structural features specific to buildings located in North Jakarta, Indonesia.
2.2. Current Use: building features and usage indications, such as signage, layout, and surrounding context specific to
buildings located in North Jakarta, Indonesia.
2.3. Storeys: the number of storeys and structural elements, such as windows and floor separations specific to buildings
located in North Jakarta, Indonesia.
3. Your goal is to infer the most likely <building_construction>, <building_current_use>, and <building_storeys>.
4. If it's not possible to infer the most likely <building_construction>, <building_current_use>, and <building_storeys>
from the main building in the centre of the image, infer from surrounding buildings with similar characteristics or
features.
</instructions>

<step_by_step_thinking>
Think before you predict the building typologies (construction, current use, and storeys) in <thinking_construction>,
<thinking_current_use>, and <thinking_storeys>.

Structured thinking guide:
1. Identify and describe features of the building relevant to the building construction, current use, and storeys.
2. Analyse the building's surroundings to add context for the building construction, current use, and storeys
prediction.
3. Infer the most suitable building construction, current use, and storeys.
4. Compare and validate the predicted building construction, current use, and storeys with the features and surrounding
environment you described earlier (correct the previous prediction, if necessary).

After comparison and validation, proceed to infer <building_construction>, <building_current_use>, and
<building_storeys> and <reasoning_construction>, <reasoning_current_use>, and <reasoning_storeys> behind it,
respectively.
</step_by_step_thinking>

<formatting>
Only select <building_construction> from this list: ["Wood", "Stone", "Brick", "Steel", "Reinforced Concrete", "Other
Metal", "Other Natural Material", "Other Man-Made Material"].
Only select <building_current_use> from this list: ["Mixed-use", "Residential", "Retail", "Industry & Businesses",
"Community Services", "Recreation & Leisure", "Transport", "Utilities & Infrastructure", "Defence", "Agriculture",
"Vacant & Derelict", "Unclassified, presumed non-residential"].
Only select <building_storeys> from this list: ["1", "2-3", "4-7", "8-10", "11-39", "40+"].

Organize your answer in the following JSON format containing 9 keys:
{
    "thoughts_construction": <thinking_construction>,
    "construction": <building_construction>,
    "reason_construction": <reasoning_construction>,
    "thoughts_currentuse": <thinking_current_use>,
    "currentuse": <building_current_use>,
    "reason_currentuse": <reasoning_current_use>,
    "thoughts_storeys": <thinking_storeys>,
    "storeys": <building_storeys>,
    "reason_storeys": <reasoning_storeys>
}

The meaning of 9 keys:
"thoughts_construction": detailed, step-by-step breakdown of your reasoning process following the 4 points in the
structured thinking guide on predicting <building_construction> (each thinking step should be at around 400 characters
or 75 tokens max).
"construction": the most likely <building_construction> chosen from the provided list.
"reason_construction": a concise explanation supporting your prediction, considering visible material characteristics,
structural features, and any other relevant details specific to buildings located in North Jakarta, Indonesia (please do
not use line breaks in the reason and answer in 400 characters or 75 tokens max).
"thoughts_currentuse": detailed, step-by-step breakdown of your reasoning process following the 4 points in the
structured thinking guide on predicting <building_current_use> (each thinking step should be at around 400 characters or
75 tokens max).
"currentuse": the most likely <building_current_use> chosen from the provided list.
"reason_currentuse": a concise explanation supporting your prediction, considering visible features, usage indications,
signage, layout, and any other relevant details specific to buildings located in North Jakarta, Indonesia (please do not
use line breaks in the reason and answer in 400 characters or 75 tokens max).
"thoughts_storeys": detailed, step-by-step breakdown of your reasoning process following the 4 points in the structured
thinking guide on predicting <building_storeys> (each thinking step should be at around 400 characters or 75 tokens
max).
"storeys": the most likely <building_storeys> chosen from the provided list.
"reason_storeys": a concise explanation supporting your prediction, considering visible storeys, structural elements
like windows and floor separations, and any other relevant details specific to buildings located in North Jakarta,
Indonesia (please do not use line breaks in the reason and answer in 400 characters or 75 tokens max).
</formatting>
```

*(d) Sample Prompts for Chain-of-Thought Merged*

*Fig A3. Prompt Strategies*